\documentclass{trbunofficial}
\usepackage{graphicx}
\usepackage{indentfirst}
\usepackage{algorithm}
\usepackage[noend]{algpseudocode}
\usepackage{amssymb}

\AuthorHeaders{\textit{Han, Filev, and {\"O}zg{\"u}ner}}
\title{An Online Evolving Framework for Modeling The Safe Autonomous Vehicle Control System via Online Recognition of Latent Risks }

\author{%
  \textbf{Teawon Han, Ph.D. Student}\\
  Department of Electrical and Computer Engineering\\
  The Ohio State University, Columbus, OH 43210, USA, \\
  Email: han.394@osu.edu\\
  \hfill\break%
  \textbf{Dimitar Filev, Ph.D.}\\
  Research \& Advanced Engineering\\
  Ford Motor Company, Dearborn, MI 48121, USA\\
  Email: dfilev@ford.com\\ 
  \hfill\break
  \textbf{{\"U}mit {\"O}zg{\"u}ner, Professor Emeritus}\\
  Department of Electrical and Computer Engineering\\
  The Ohio State University, Columbus, OH 43210, USA, \\
  Email: ozguner.1@osu.edu
}

\begin{document}
\maketitle
\section{Abstract}
An online evolving framework is proposed to support modeling the safe Automated Vehicle (AV) control system by making the controller able to recognize unexpected situations and react appropriately by choosing a better action. Within the framework, the evolving Finite State Machine (e-FSM), which is an online model able to (1) determine states uniquely as needed, (2) recognize states, and (3) identify state-transitions, is introduced. 

In this study, the e-FSM's capabilities are explained and illustrated by simulating a simple car-following scenario. As a vehicle controller, the Intelligent Driver Model (IDM) is implemented, and different sets of IDM parameters are assigned to the following vehicle for simulating various situations (including the collision). While simulating the car-following scenario, e-FSM recognizes and determines the states and identifies the transition matrices by suggested methods.

To verify if e-FSM can recognize and determine states uniquely, we analyze whether the same state is recognized under the identical situation. The difference between probability distributions of predicted and recognized states is measured by the Jensen-Shannon divergence (JSD) method to validate the accuracy of identified transition-matrices. As shown in the results, the Dead-End state which has latent-risk of the collision is uniquely determined and consistently recognized. Also, the probability distributions of the predicted state are significantly similar to the recognized state, declaring that the state-transitions are precisely identified.


\hfill\break%
\noindent\textit{Keywords}: Decision Making Process, Markov Chain, Finite State Machine, Automated Vehicle, Latent Risk Detection 
\newpage

\section{1. Introduction}
\indent
For the intelligent control system, it is required to accurately perceive the current state to react appropriately based on given criteria. For example, the automated vehicle (AV) controller needs to recognize the traffic situations precisely for the decision making of maneuver operation. However, it is a challenge to design the decision-making framework for AV due to unanticipated and complex traffic situations. 


For many decades, rule-based, supervised learning, and unsupervised learning approaches have been proposed as a decision-making method to control the automated vehicle (AV) in different manners and levels. As a rule-based approach, the Hierarchical Finite State Machine (HFSM) has been implemented in the AV control framework. In the 2007 DARPA urban challenge, the Hybrid State System (HSS) was proposed to control the AV (OSU-ACT) as shown in \cite{kurt2008hybrid, redmill2008ohio}. The HSS consists of a Discrete State System (DSS), a Continuous State System (CSS), and an Interface layer. The DSS consists of HFSM for the high-level decision-making while the CSS maneuvers the AV at a low-level. In \cite{liu2007human}, HFSM is implemented to create a decision-making module by analyzing the human driver's behaviors for intersection driving. The module is embedded in HSS to estimate the human driver's decision in \cite{gadepally2013framework}. For a set of driving decisions in merging into a convoy, \cite{kurt2011probabilistic} proposes to use the transition probability in FSM instead of conditions of state-transition. For automated driving on the highway, the driving strategy decision model which consists of 2-levels HFSM is proposed in \cite{noh2017decision}, and \cite{zhang2017finite} suggests the FSM based automated driving controller with a stochastic gradient optimization method.

As a supervised-learning approach, deep neural-network (DNN) has been proposed as the AV controller. \cite{al2017deep} is implemented the GoogLeNet to obtain accurate affordance parameters that are used to determine the optimal control actions. The end-to-end learning which is mapping the camera images with optimal controls via a convolutional neural network (CNN) is claimed by \cite{bojarski2016end} for self-driving cars. As the unsupervised-learning approach, the reinforcement-learning (RL) with the multilayer perceptron (MLP) is proposed for automated speed and lane-change decision making in \cite{hoel2018automated}. The Q-learning algorithm is implemented to learn the optimal policies for various driving behaviors in \cite{you2018highway}. Also, \cite{hejase2018identification} introduced how to analyze the latent-risks by using a Backtracking Process Algorithm (BPA).

In spite of the fact that previously proposed decision-making methodologies for AV control system derive optimal controls in several scenarios, they have some limitations. Rule-based and supervised-learning methods cannot recognize unexpected situations so that the AV controller cannot react appropriately under unknown circumstances. In other words, the performance of rule or supervised-learning based decision-making is guaranteed only under initially anticipated situations. Through the unsupervised-learning method, it is possible to learn the best action under newly encountered states, but its decision-making process depends on a pre-designed reward function and the current state. Its limited capabilities bring the case: the best action at the current state could be worse for the future state. 

To surpass the limitations, a combination of the evolving clustering method and Markov Chain, which is suggested in \cite{filev2013generalized}, is implemented to derive an evolving module named evolving Finite State Machine (e-FSM). The e-FSM is proposed for online state determination and recognition with identification of state-transitions. Also, an online evolving framework is introduced to show how e-FSM helps the AV controller to choose a better action regardless of the AV controller's type. The rest of the paper is organized as follows: Section 2 describes how e-FSM determines and recognizes states identifying the state-transitions. In Section 3, e-FSM's capabilities are validated via analysis of experimental results. An online evolving framework for the safe AV controller system is introduced in Section 4. Finally, the contributions of this study are summarized in Section 5.  

\section{2. evolving finite state machine Model} \label{sec:section2}
\begin{figure}[!h]
  \centering
  \includegraphics[width=1\textwidth]{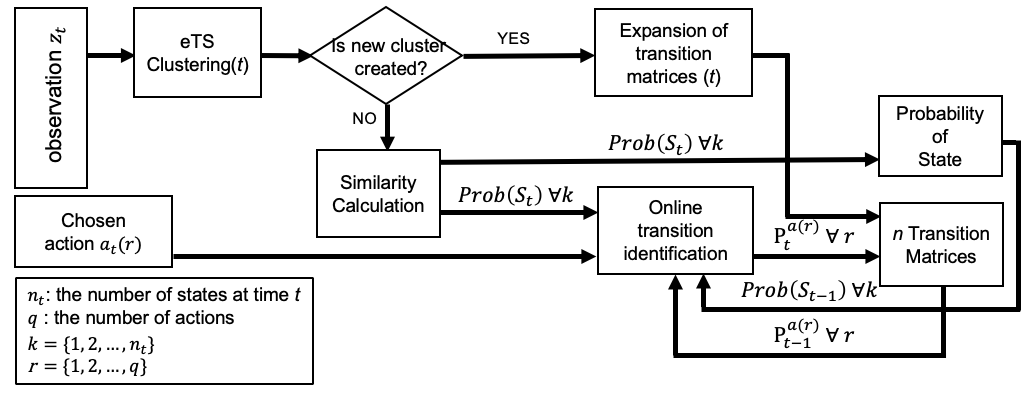}
  \caption{A framework of evolving Finite State Machine}
  \label{fig:eFSM_framework}
\end{figure}

The fundamental concept of e-FSM is inspired by the general finite state machine (FSM), which consists of states (nodes) and transition-conditions (links). A framework of e-FSM is proposed as shown in \textbf{Figure \ref{fig:eFSM_framework}}. e-FSM consists of states and conditional transition probability, but it can evolve its structure by creating new states via clustering observations $z_t$ over time. Also, state-transitions are represented by multiple matrices, and each matrix is correlated with each action. The matrices are identified or expanded as the following: once a new state is created, the dimension of all transition matrices is expanded. Otherwise, one of the transition matrices, which is correlated with a chosen action, is identified. The specific properties and configurations of state and transition matrices are discussed in the following sub-sections.

\subsection{2.1. The properties of e-FSM}
e-FSM has a set of states $S_t$ at time $t$, a set of actions $A$, and conditional transition probabilities in the matrix form $\mathbf{P}_t^{a(r)}$ where $a(r)$ is a chosen action at time $t$ and $1\leq r\leq q$; $q$ is the total number of actions. As expected in the notations, the state set and transition matrices can be changed over time, whereas the action set is fixed. 

\subsubsection{2.1.1. State}
The $state$ $set$ is initially empty, $S_0=\{\}$, but is getting to have the various number of states over time such that $S_{t\geq 1}=\{s_t(1), s_t(2), ..., s_t(n_t)\}$ where $n_t$ is the total number of states at time $t$. Due to that the number of states is not fixed, $n_t$ has the following properties: $n_0=0$, otherwise $1\leq n_{t}\leq n_{t+1}\leq n_{\infty}\leq \infty$. Each state $s_t(\cdot)$ is represented by a center of clustered observations at time $t$, referring to an unique situation. The observation $z_t$ consists of $m$ number of variables which are continuous and/or discrete type of information. For instance, the observation can be defined to represent driving situations such as $z_t=\left[headway_t(\alpha), v_t(\alpha), pos_t(\alpha)\right]^T$ of which elements refer to a distance between vehicle $\alpha$ ($veh_{\alpha}$) and its preceding vehicle ($veh_p$), velocity of $veh_{\alpha}$, and position of $veh_{\alpha}$. The recognized state at time $t$ is represented by the probability distributions such that $Prob(S_t)=\left[Prob(s_t(1)), Prob(s_t(2)), ..., Prob(s_t(n_t))\right]^T$. The detail steps to calculate the probability distributions are discussed in Section 2.2.

\subsubsection{2.1.2. Action}
The $action$ $set$ needs to be set by the fixed number of finite discrete actions such that $A_d=\{a(1), a(2), ..., a(q)\}$ where $q$ is the total number of actions and $q\geq 1$. The discrete action set can be obtained by encoding the continuous action set with an arbitrary chosen interval $\delta$. For example, the continuous longitudinal acceleration set $A_{c}=(-1.0 \,\,\,\,\,1.0] [m/s]$ can be encoded to the discrete action set with $\delta=0.5$ such that $A_{d}=\{a(1), \,\,a(2), \,\,a(3), \,\,a(4)\}$ where $a(1)=(-1.0 \,\,\,\,-0.5],\,\,\,a(2)=(-0.5 \,\,\,\,0.0],\,\,\,a(3)=(0.0 \,\,\,\,0.5],$ and $a(4)=(0.5 \,\,\,\,\,\,\, 1.0]$. The controller chooses an optimal action in the action-set, and the chosen action is used to identify transition matrices in e-FSM.

\subsubsection{2.1.3. State-Transitions}
The transitions among determined states are represented by conditional probabilities which are illustrated as transition matrices. Each transition matrix is correlated with each action, therefore, the total number of transition matrices is identical to the total number of actions. Due to that the action set is fixed, the number of transition matrices is not changed, but their dimensions are varied over time based on the determination of new states. Given a chosen action $a(r) \in A_d$, the transition probability from state $s_t(i)$ to $s_t(j)$ is determined such as $P_t^{a(r)}(i,j)=Prob\left(state_{next}=s_t(j) | state_{prior}=s_t(i), a_t=a(r)\right)$. The q number of transition matrices are defined such that $\mathbf{P}_t^{a(r)}=\{P_t^{a(r)}(i,j)\}$ where $r=\{1, 2, ..., q\}$, $1 \leq i,j \leq n_t$, and $dim\left(\mathbf{P}_t^{a(r)}\right)\forall r=R^{n_t \times n_t}$ at time $t\geq1$. The details for identification and expansion of transition matrices are discussed in Section 2.3.

\subsubsection{2.1.4. Example of e-FSM's evolving sequence}
\begin{figure}[!h]
  \centering
  \includegraphics[width=1.0\textwidth]{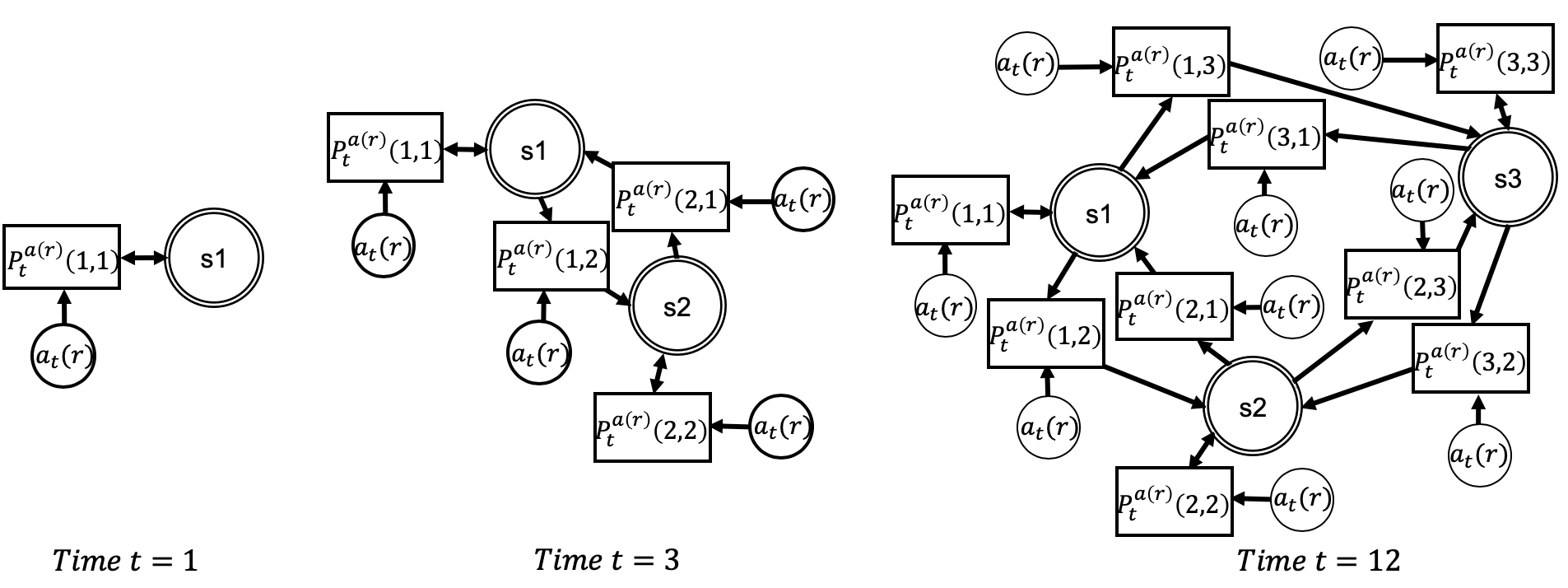}
  \caption{An example of e-FSM's evolving sequences}
  \label{fig:eFSM_example}
\end{figure}

An example is illustrated in \textbf{Figure \ref{fig:eFSM_example}} to explain how e-FSM is evolving over time.  $s1, s2,$ and $s3$ are states, $a(r)$ is a chosen action by the controller at time $t$, and $P_t^{a(r)}(i,j)$ refers the transition probability from state $i$ to $j$ given a chosen action $a(r)$. After state $s1$ is determined at time $t=1$, no state is additionally determined until $t=3$, therefore, only the transition probability from $s1$ to $s1$ are identified based on the chosen actions. At time $t=3$, a new state $s2$ is determined so that the dimension of all transition matrices is expanded from $R$ to $R^{2 \times 2}$. Then, four state-transitions are identified until $t=12$ when a new state $s3$ is created. 

\subsection{2.2. The principle of the online state determination and recognition}

Once a set of variables (or observations) is set for representing the state, e-FSM determines or recognizes the states. For the determination of the states, evolving Takagi-Sugeno (eTS) \cite{angelov2004approach, filev2011real} which is one of the online-clustering methods is implemented. This is because that eTS is able to regulate whether the input-data (observation) defined such as \textbf{Equation \ref{def:obs}} is grouped into one of the existing clusters or becomes a center of the new cluster. This feature can be directly applicable to the state determination considering each cluster as a state. 

\subsubsection{2.2.1. Determination of the new states via eTS online clustering method}
The eTS consists of three steps as the follows: First, it calculates a potential of the input-data (single or multiple variables) $z_t$ by \textbf{Equation \ref{eq:eTS1}}. Second, the potentials of all existing cluster centers $z_t^{*i}$ are updated by \textbf{Equation \ref{eq:eTS2}}, where $z_t^{*i}$ is a center of $i^{th}$ cluster. Lastly, it decides whether the input-data $z_t$ should be classified to one of the existing clusters or be a center of the new cluster by considering the following two conditions: 

\noindent \textbf{Condition\,1}: $\tilde{P}_{t}\left( z_{t} \right) > max\left[ \tilde{P}_{t}\left( z_{t}^{*i}\right) \forall i \right]$, which means that a potential of current input-data $\tilde{P}_t(z_t)$ is greater than the potentials of all existing cluster centers $\tilde{P}_t(z_t^{*i}) \forall i, i=\{1, ..., R\}$ and $R$ is the total number of existing clusters. 

\noindent \textbf{Condition\,2}: $ \left\| z_{t} - z_{t}^{*\omega} \right\|<\epsilon $, which means that the minimum euclidean distance between the input-data $z_t$ and the closest cluster center $z_t^{*\omega}$ is less than $\epsilon$, where $\omega=\underset{i}{argmin}\left\| z_{t} - z_{t}^{*i}\right\| \forall i$. 
\hfill\break%
If $Condition\:1$ and $Condition\:2$ are satisfied, then one of the existing cluster center  $z_t^{*\omega}$ which is the closest to $z_t$ is replaced by $z_t$. If only $Condition\:1$ is satisfied, then a new cluster centered $z_t$ is created. $b_1$ is initialized by $0$, and the first input-data is set as a center of the new cluster having $\tilde{P}_1(z_1)=1$ instead of the potential calculation by \textbf{Equation \ref{eq:eTS1}}. $\epsilon$ and $\rho$ are arbitrarily assigned, which affects the frequency of the new cluster creation. 

\begin{equation}
  \label{def:obs}
    z_t=[var(1), var(2), ..., var(m)]
\end{equation}
\begin{flalign}
    \label{eq:eTS1}
    &\tilde{P}_{t}\left( z_{t}\right)=\frac{t-1}{(t-1)(a_{t}+1)-2c_{t}+b_{t}}
\end{flalign}
where 
\begin{flalign}
    \label{eq:eTS1_1}
    \:\:\:\:\:\:\:\:\:\:\:\:\:\:\:&a_{t}=z_{t}^{T}z_{t}\\
    &b_{t}=b_{t-1}+z_{t-1}^{T}z_{t-1}\\
    &c_{t}=\sum_{j=1}^{m}\left( z_{t}^{j}\sum_{k=1}^{t-1}z_{k}^{j}\right) ; m=dim(z_t)
\end{flalign}
\begin{flalign}
    \label{eq:eTS2}
    &\tilde{P}_{t}(z_{t}^{*i})=\frac{(t-1)\tilde{P}_{t-1}(z_{t}^{*i})}{(t-2)+\tilde{P}_{t-1}(z_{t}^{*i})\left[1+\rho\left\|z_{t}^{*i}-z_{t-1}\right\|^2\right]}
\end{flalign}

\subsubsection{2.2.2. Recognition of the current state}
The e-FSM recognizes the current state based on observation $z_t$ and existing states $S_t$. The recognition of the current state is represented by the probability distributions over the existing states. When a new state is not created by eTS, the similarity function defined as \textbf{Equation \ref{eq:similarity}} is called to calculate how much $z_t$ is similar to the existing states $s_t(i) \in S_t$, $i=\{1,\,\,2,\,\,...,\,\, n_t\}$. Due to that each similarity $\eta_{t}^{i}(z_{t})$ is normalized, it is bounded 0 to 1, therefore, the probability distributions of the state at time $t$, $Prob(S_t)$, is defined in e-FSM as shown in \textbf{Equation \ref{eq:prob_state1}} and \textbf{Equation \ref{eq:prob_state2}}.
\begin{flalign}
    \label{eq:similarity}
    &\lambda_{t}^{i}(z_{t})=\frac{\eta_{t}^{i}(z_{t})}{\sum_{j}\eta_{t}^{j}(z_{t})} \text{\:\:\:\:where  \:\: } \eta_{t}^{i}(z_{t})=exp\left(-\frac{(z_{t}-z_{t}^{*i})^{T}(z_{t}-z_{t}^{*i})}{var(z_{t}^{*i})}\right)\\
    \label{eq:prob_state1}
    &Prob(S_t)=[Prob(s_t(1)), Prob(s_t(2)), ..., Prob(s_t(n_t))]^T\\
    \label{eq:prob_state2}
    &Prob(s_t(k))=\lambda_t^i(z_t)\,\,\text{ where }\, 1 \leq i\leq n_t
\end{flalign}

\subsection{2.3. The principle of the online state-transition identification}
The state-transition identification is critical for e-FSM to predict future states. Due to that transition matrices of which each is correlated with each action are implemented, it is possible to realize what kind of the state-transitions will be appeared based on a chosen action. 
The stochastic method is proposed to identify the transition-matrices, but also the logic is introduced to expand the dimension of transition-matrices because the number of states is not fixed increasing over time as needed in e-FSM.

\subsubsection{2.3.1. Online identification method for a transition matrix in Markov Chain}
In Markov Chain, a transition probability from $state$ $s(i)$ to $s(j)$ denoting $\pi_{i,j}$ is defined by \textbf{Equation \ref{eq:transition_markov_model1}} and \textbf{\ref{eq:transition_markov_model2}}, where $f_{ij}(t)=1$ when state-transition from $s(i)$ to $s(j)$ is observed at time-step $t$, and $f_{i}(t)=1$ when state-transition is initiated from $s(i)$ at time-step $t$ is observed. On top of the transition probability definition in Markov Chain, the online state-transition identification method as shown in \textbf{Equation \ref{eq:transition_matrix}-\ref{eq:online_identify2}} is proposed by \cite{filev2013generalized} to implement Markov models for real-time modeling of continuous systems. In the equations, $\varphi$ is a learning rate, $\tau(t)$ and $\gamma(t)$ are probability distributions of states at time step $t-1$ and $t$ respectively, and $1_N$ is N-dimensional ones-vector where $N$ is the total number of states. For initialization of a transition matrix, $F(0)$ and $F_o(0)$ are set by $\bar{\varepsilon}E$ and $F(0)\cdot 1_N$ respectively; $\bar{\varepsilon}$ is a small non-negative constant for avoiding singularity; $E$ is a compatible-size $(N \times N)$ matrix having unit elements.     
\begin{flalign}
    \label{eq:transition_markov_model1}
    &N_{ij}(k)=\sum_{t=1}^{k}f_{ij}(t), \,\,\,\, N_{oi}(k)=\sum_{t=1}^{k}f_{i}(t)\\
    \label{eq:transition_markov_model2}
    &\pi_{i,j}(k)\approx\frac{N_{ij}(k)}{N_{oi}(k)}=\frac{F_{ij}(k)}{F_{oi}(k)}, \text{ where } F_{ij}(k)= \frac{N_{ij}}{k},\,\,\,\,F_{oi}(k)=\frac{N_{oi}(k)}{k}\\
    \label{eq:transition_matrix}
        &\Pi(k)=diag(F_0(k))^{-1}F(k)\\
    \label{eq:online_identify1}
        &F(k)=F(k-1)+\varphi(\tau(k)\gamma(k)^T-F(k-1))\\
    \label{eq:online_identify2}
        &F_o(k)=F_o(k-1)+\varphi(\tau(k)\gamma(k)^T1_N-F_o(k-1))
\end{flalign}
\subsubsection{2.3.2. The online identification and expansion methods for transition matrices in e-FSM}
The online state-transition identification method proposed in \cite{filev2013generalized} is implemented in e-FSM with some modifications. Due to that the new states are determined from time to time in e-FSM, the dimension of transition matrices should be expanded to represent state-transitions between all existing states. 
Recalling the notation $n_t$ meaning the total number of determined states by time $t$, the dimension of the transition matrices should always be $R^{n_t \times n_t}$.

Because multiple transition matrices are implemented in e-FSM for representing state-transitions based on the chosen actions, \textbf{Equation \ref{eq:transition_matrix}}, \textbf{\ref{eq:online_identify1}}, and \textbf{\ref{eq:online_identify2}} are re-defined by \textbf{Equation \ref{eq:transition_matrix_efsm}}, \textbf{\ref{eq:online_identify1_efsm}}, and \textbf{\ref{eq:online_identify2_efsm}}, where $r=\{1,2, ..., q\},$ $ \tau(t)=Prob(S_{t-1})$, and $\gamma(t)=Prob(S_t)$. For the initialization, $F^{a(r)}(0) \forall r$ is set by $\bar{\varepsilon}$ rather than $\bar{\varepsilon}E$. This is because the number of states in e-FSM is not fixed but is varied over time. Only one of the transition matrices which is correlated with a chosen action is identified by \textbf{Equation \ref{eq:transition_matrix_efsm}} at a time, and others are kept without any changes.

\begin{flalign}
    \label{eq:transition_matrix_efsm}
        &\mathbf{P}_t^{a(r)}=diag(F_0^{a(r)}(t))^{-1}F^{a(r)}(t)\\
    \label{eq:online_identify1_efsm}
        &F^{a(r)}(t)=F^{a(r)}(t-1)+\varphi(\tau(t)\gamma(t)^T-F^{a(r)}(t-1))\\
    \label{eq:online_identify2_efsm}
        &F_o^{a(r)}(t)=F_o^{a(r)}(t-1)+\varphi(\tau(t)\gamma(t)^T1_N-F_o^{a(r)}(t-1))
\end{flalign}
\hfill\break%
As shown in \textbf{Figure \ref{fig:eFSM_framework}}, when a new state (cluster) is created, the expansion of transition matrices is executed following two steps. First step is simply inserting a new row and column into $F^{a(r)}$ matrix for all $r$ so that $dim(F^{a(r)}(t))\forall r$ become $R^{n_t \times n_t}$ where $n_t=n_{t-1}+1$. Then, the elements in the new row and column are initialized by $\bar{\varepsilon}$. Second step is the update of $F_o^{a(r)}$ vector for all $r$ by adding a row so that $dim(F_o^{a(r)}(t))\forall r$ is increased from $R^{n_{t-1}}$ to $R^{n_t}$. In $F_o^{a(r)}(t)$ $\forall r$, $\bar{\varepsilon}$ is added to first $n_{t-1}$ elements and the last element is initialized by $n_t\bar{\varepsilon}$. 

For instance, assuming the action set consists of two actions such that $A=\{a(1), a(2)\}$ in the given example \textbf{(Figure \ref{fig:eFSM_example})}, two transition matrices at time $t=11$, $\mathbf{P}_{11}^{a(1)}$ and $\mathbf{P}_{11}^{a(2)}$, can be calculated by using \textbf{Equation \ref{eq:transition_matrix_efsm}, \ref{eq:online_identify1_efsm}} and \textbf{\ref{eq:online_identify2_efsm}}, where \begin{center}
$F^{a(1)}(11)=\begin{bmatrix}
 \hat{f}_{1,1}^{ a(1)}(11) & \hat{f}_{1,2}^{ a(1)}(11) \\ 
 \hat{f}_{2,1}^{ a(1)}(11) & \hat{f}_{2,2}^{ a(1)}(11)
\end{bmatrix}$, \quad $F^{a(2)}(11)=\begin{bmatrix}
 \hat{f}_{1,1}^{ a(2)}(11) & \hat{f}_{1,2}^{ a(2)}(11) \\ 
 \hat{f}_{2,1}^{ a(2)}(11) & \hat{f}_{2,2}^{ a(2)}(11)
\end{bmatrix}$, \\ $F_o^{a(1)}(11)=[\hat{f}_1^{a(1)}(11), \hat{f}_2^{a(1)}(11)]^T$, \quad $F_o^{a(2)}(11)=[\hat{f}_1^{a(2)}(11), \hat{f}_2^{a(2)}(11)]^T.$\end{center} 
A new state $s3$ is determined at time $t=12$, therefore, the dimension of the two transition matrices needs to be expanded. First, the matrices, $F^{a(1)}(11)$ and $F^{a(2)}(11)$, are expanded and initialized such that $F^{a(1)}(12)=\begin{bmatrix}
 \hat{f}_{1,1}^{ a(1)}(11) & \hat{f}_{1,2}^{ a(1)}(11) & \bar{\varepsilon}\\ 
 \hat{f}_{2,1}^{ a(1)}(11) & \hat{f}_{2,2}^{ a(1)}(11) & \bar{\varepsilon}\\
 \bar{\varepsilon} & \bar{\varepsilon} & \bar{\varepsilon}
\end{bmatrix}$ and $F^{a(2)}(12)=\begin{bmatrix}
 \hat{f}_{1,1}^{ a(2)}(11) & \hat{f}_{1,2}^{ a(2)}(11) & \bar{\varepsilon}\\ 
 \hat{f}_{2,1}^{ a(2)}(11) & \hat{f}_{2,2}^{ a(2)}(11) & \bar{\varepsilon}\\
 \bar{\varepsilon} & \bar{\varepsilon} & \bar{\varepsilon}
\end{bmatrix}$. Second, the vectors, $F_o^{a(1)}(11)$ and $F_o^{a(2)}(11)$, are updated such that $F_o^{a(1)}(12)=[\hat{f}_1^{a(1)}(11)+\bar{\varepsilon}, \hat{f}_2^{a(1)}(11)+\bar{\varepsilon}, 3\times\bar{\varepsilon}]^T$, and $F_o^{a(2)}(12)=[\hat{f}_1^{a(2)}(11)+\bar{\varepsilon}, \hat{f}_2^{a(2)}(11)+\bar{\varepsilon}, 3\times\bar{\varepsilon}]^T$.

To assist the controller's decision making, e-FSM can provide the controller what state transitions will occur in the future based on chosen actions by calculating the probability distributions of the future state given each possible action. Given $a(r)$, $\mathbf{P}_t^{a(r)}$, and $Prob(S_t)$, the probability distributions of the state at $t+1$ and $t+k\,\,(k\geq2)$ can be obtained by \textbf{Equation \ref{eq:ProbDist_future}} and \textbf{\ref{eq:ProbDist_future_multi}} respectively.

\begin{flalign}
    \label{eq:ProbDist_future}
        &Prob_{pred}(S_{t+1})=\mathbf{P}_t^{a(r)}\cdot Prob(S_t) \\
    \label{eq:ProbDist_future_multi}
        &Prob_{pred}(S_{t+k})=({\mathbf{P}_t^*})^{k-1}\cdot Prob_{pred}(S_{t+1}) 
\end{flalign}
\hfill\break%
where $\mathbf{P}_t^*=\{P_t^*(i,j)\},\,\,1\leq i,j \leq n_t$, and $P_t^*(i,j)$ is the marginal probability of the state transition from $i$ to $j$ defining by \textbf{Equation \ref{eq:marginal}}; the uniform distribution is applied to the probability of action such that $Prob(a_t=a(r))=1/q,\,\,r=\{1,\,\,2,\,\,...,\,\, q\}$ in the equation.
\begin{equation}
  \label{eq:marginal}
    P_t^*(i,j)=\sum_{r=1}^{q}Prob\left(state_{next}=s_t(j) | state_{prior}=s_t(i), a_t=a(r)\right)\cdot Prob(a_t=a(r))
\end{equation}


\section{3. Experimental Setting and Results}
Using the Simulation of Urban Mobility (SUMO), a car-following scenario is simulated to show whether the states are uniquely determined and recognized and to show how accurate the state-transitions are identified through e-FSM. 

\subsection{3.1. Experimental Settings}
\begin{figure}[!h]
  \centering
  \includegraphics[width=0.9\textwidth]{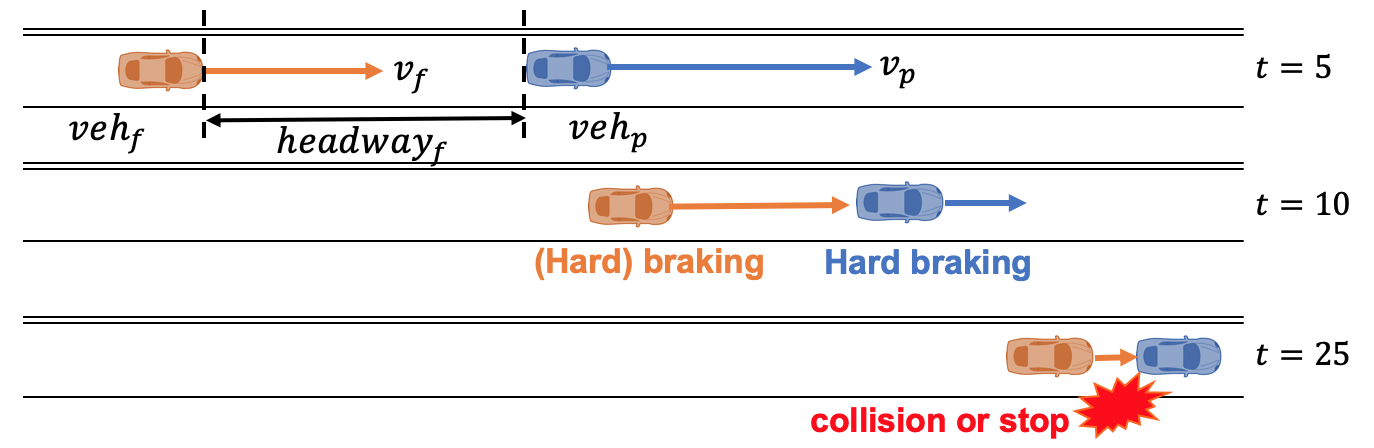}
  \caption{The scenario for the experiment}\label{fig:scenario}
\end{figure}

There is a moment that the driver does not have eligible actions to avoid accidents. For instance, while the following vehicle is driving with the short safe-distance, if the preceding vehicle takes a full-brake, there is no chance for the following vehicle to avoid a collision. Calling the inevitable collision state(s) by the Dead-End (DE) state(s), a simple car-following scenario is designed 
to make the DE state happen on the one-way road based on the following vehicle's speed control as shown in \textbf{Figure \ref{fig:scenario}}. The specific scenario settings are determined as the following: while the following and preceding vehicles, $veh_f$ and $veh_p$, are driving on the one-way road, each vehicle's speed, $v_f$ and $v_p$, is controlled by the individual controller. When the preceding vehicle's speed reaches its max-speed $v_p^{max}$, it will take a full brake to stop like as the emergency-stop. The car-following scenario is terminated either when a collision is observed or after 35 seconds of simulation; the unit-time is 0.01 secs; therefore, 3500 steps are simulated for a single scenario.

The Intelligent Driver Model (IDM) which is a microscopic car-following model is implemented as the controller of the both vehicles because the model is designed and validated to create the realistic longitudinal car-following motions as discussed in \cite{kesting2010enhanced, treiber2000congested}. The IDM (\textbf{Equation \ref{eq:IDM1}} and \textbf{\ref{eq:IDM2}}) consists several parameters of which some are observations ($v_{\alpha}, \Delta v_{\alpha}, s_{\alpha})$, others ($a^{(\alpha)}, v_0^{(\alpha)}, s_0^{(\alpha)}, T^{(\alpha)}, b^{(\alpha)}$) are  driver's preferences of the vehicle $\alpha$; the desired maximum acceleration, desired velocity, desired headway, desired time-headway, and desired maximum deceleration are represented by $a^{(\alpha)}$, $v_0^{(\alpha)}$, $s_0^{(\alpha)}$, $T^{(\alpha)},$ and $b^{(\alpha)}$ respectively; $\Delta v_{\alpha}=v_{\alpha}-v_p$; $s_{\alpha}$ refers a current headway of $veh_{\alpha}$. By assigning different sets of IDM parameters, distinct driving styles of the controller which choose different actions (e.g., longitudinal acceleration in the scenario) under the identical situation are obtained like as more aggressive or normal type of the longitudinal speed controller. Two different sets of IDM parameters are pre-determined and used to obtain the different types of following vehicle, and a set of IDM parameters is assigned to the preceding vehicle. Therefore, the following vehicle encounters identical situations, but it reacts differently based on its controller type. The parameter set, $\left[a^{(\alpha)}\,_{[m/s^2]},\,\,v_0^{(\alpha)}\,_{[m/s]},\,\,s_0^{(\alpha)}\,_{[m]},\,\,T^{(\alpha)}\,_{[sec]},\,\,b^{(\alpha)}\,_{[m/s^2]}\right]$, is set by $\left[1.2,\,\,25,\,\,1.0,\,\,1.0,\,\,2.5\right]$, $\left[2.25,\,\,28,\,\,0.8,\,\,0.3,\,\,2.0 \right]$, and $\left[1.25,\,\,25,\,\,2.0,\,\,1.5,\,\,2.0\right]$ for the preceding vehicle, the aggressive following vehicle, and the normal type following vehicle respectively.

\begin{flalign}
    \label{eq:IDM1}
        &\dot{v_{\alpha}}=a^{(\alpha)}\left[1-\left(\frac{v_{\alpha}}{v_0^{(\alpha)}} \right)^{\delta}-\left(\frac{s^*(v_{\alpha},\Delta v_{\alpha}}{s_{\alpha}}\right)^2 \right]\\
    \label{eq:IDM2}
        &s^*(v_{\alpha},\Delta v_{\alpha})=s_0^{(\alpha)}+v_{\alpha}T^{(\alpha)}+\frac{v_{\alpha}\Delta v_{\alpha}}{2\sqrt{a^{(\alpha)} b^{(\alpha)}}}
\end{flalign}
\hfill\break%
\indent Simulating the car-following scenario, e-FSM determines and recognizes states with identifying transition matrices. The observation is defined by $z_t=[s_t(veh_f), v_t(veh_f), v_t(veh_p)]$ to represent the state in e-FSM, $\rho$ and $\epsilon$, the variables of eTS, are set by 0.85 and 0.3 respectively, the initial speed of both vehicles is set by 0, and the continuous action set (longitudinal acceleration) $A_{c}=[-2.5\,\,\,\,2.5]\,\,\,\,[m/s^2]$ is encoded by the range $0.3 m/s^2$ to the discrete action set which consists of 17 intervals such that $A_{d}=\{a(1), ..., a(17)\}$, where $a(1)=[-2.5\,\,\,-2.2),$ $a(2)=[-2.2,\,\,\,-1.9),$ $...$, $a(16)=[1.9\,\,\,2.2)$, $a(17)=[2.2\,\,\,2.5]$. 

\subsection{3.2. Experiment Process and Results}
In the car-following scenario, four cases are simulated by assigning different types of the following vehicle ($veh_f$) controller such that: (case 1) $veh_f$'s controller is set by the aggressive type, (case 2) $veh_f$'s controller is set by the normal type, (case 3) $veh_f$'s controller is initially set by the aggressive type then it is changed to the normal type at $t=15$, and (case 4) $veh_f$'s controller is initially set by the normal type then it is changed to the aggressive type at $t=10$. While simulating each case 20 times repeatedly (80 times in total), e-FSM determines or recognizes states, but also expands or identifies the transition matrices.

\begin{figure}[!h]
  \centering
  \includegraphics[width=1\textwidth]{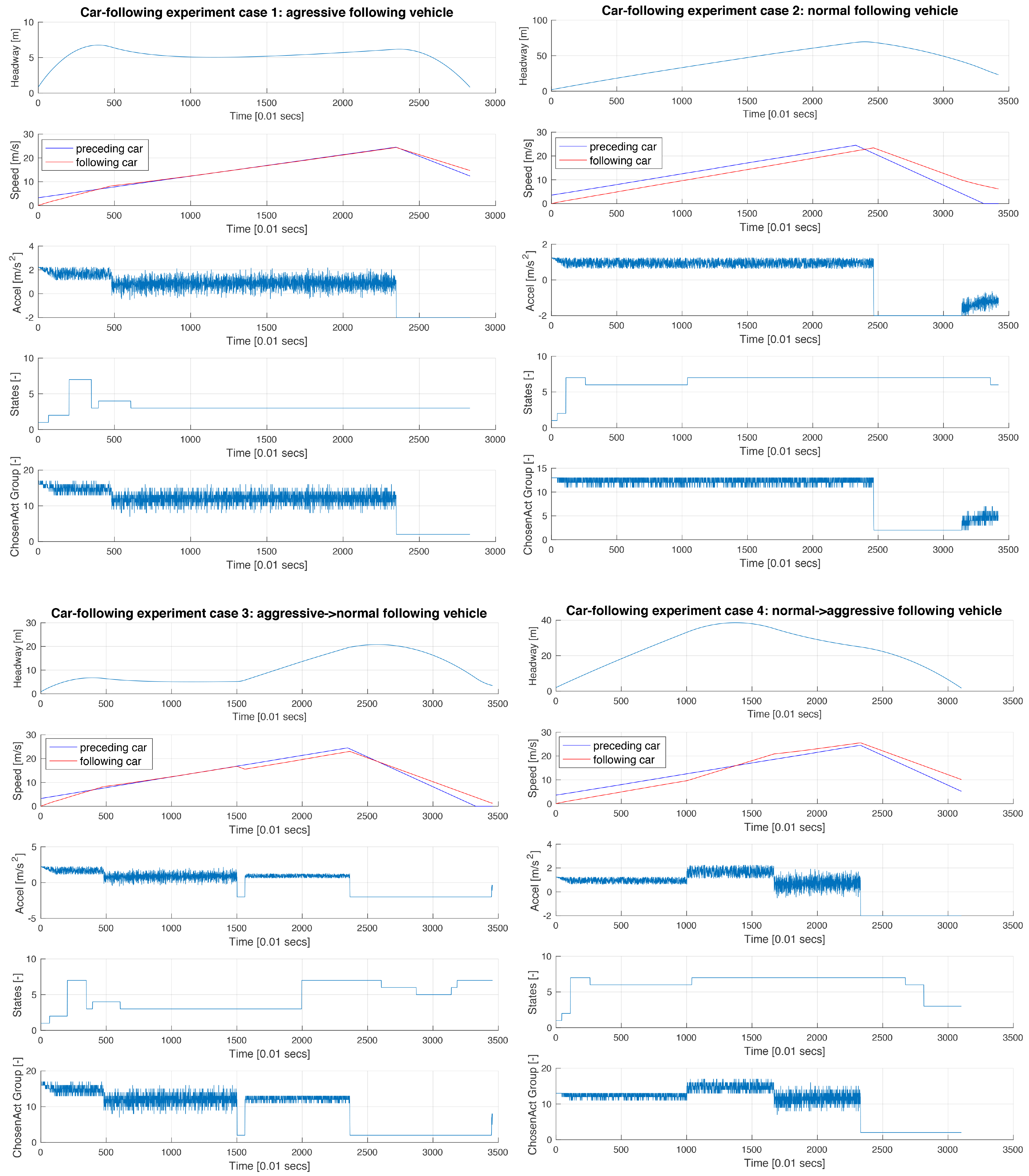}
  \caption{e-FSM's state determination and recognition in the car-following scenario with the four different following vehicle controller settings: the collision occurs in case 1 and 4.}\label{fig:result1}
\end{figure}

It is observed that the number of uniquely determined states is increased 0 to 7 after four times of simulation, then no more state is additionally determined in 80 simulations. The \textbf{Figure \ref{fig:result1}} shows the e-FSM's state recognition results in simulating the car-following scenario with the four different settings of the following vehicle's controller. In the figure, headway, speed, acceleration (the chosen continuous actions by IDM controller), the index of the recognized states by e-FSM, and the index of the interval-encoded chosen actions are shown.
\subsubsection{3.2.1. The analysis of e-FSM's state determination and recognition capabilities}

It is focused on whether the DE state is uniquely determined and consistently recognized via e-FSM in the four cases rather than inspecting each state represents what unique situations. This is because that the DE state is recognizable by observing a collision so that the e-FSM's capability can be validated by analyzing whether an identical state is recognized whenever a collision occurs.  
As shown in the results, a collision occurs at the end of simulating-horizon in case 1 and 4, whereas the following vehicle stops and goes without a collision in case 2 and 3. Also, e-FSM always recognizes the state \#3 when the collision is observed during 80 simulations without exception. In addition, it is studied whether a collision occurs or not when the following vehicle's state is recognized as the DE state (state \#3) regardless of the preceding vehicle's speed. Additional cases are set such that the preceding vehicle's full-braking is initiated at $t=11$ and $t=15$ when the following vehicle's state is recognized as a state \#3, but the preceding vehicle doesn't reach its max-speed. It is identical to case 1 except the initiating moment of the preceding vehicle's emergency stop. The collision is observed in the both cases. Therefore, it is certified that the DE state is uniquely determined by the state \#3 and consistently recognized through e-FSM. 

Changing the type of the following vehicle's controller in the middle of simulating-horizon derives state-transitions as shown in case 3 and 4. The recognized state is changed from the DE state to others after the controller's type is changed from aggressive to normal in case 3, and vice versa in case 4. It is observed that the collision can be prevented by choosing a better action in advance. In Section 4, the evolving framework is introduced to show how e-FSM can assist the AV controller's decision-making by providing the recognized latent-risks in advance. 



\begin{figure}[!h]
  \centering
  \includegraphics[width=0.85\textwidth]{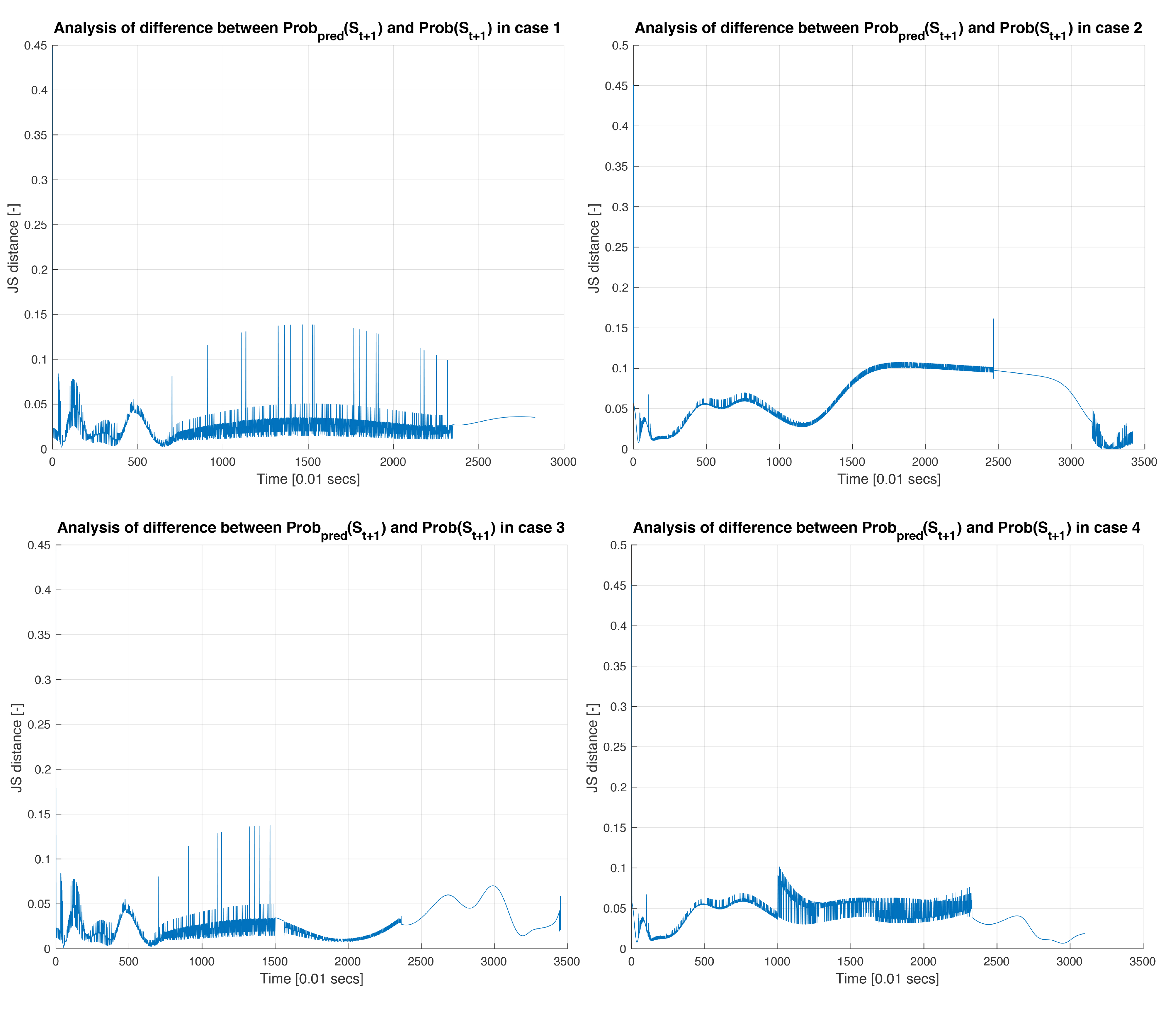}
  \caption{Difference between probability distributions of predicted and recognized states.}\label{fig:result2}
\end{figure} 

\subsubsection{3.2.2. The analysis of e-FSM's state-transition identification capability}

In e-FSM, the transition matrices are expanded or identified based on the determination of a new state or the observation of state-transitions. To show how accurate the transition matrices are identified through proposed methods, the probability distributions of predicted and recognized state, $Prob_{pred}(S_{t+1})$ and  $Prob(S_{t+1})$, are compared. The $Prob(S_{t})$ and $a(r)$ are known at every time $t$, therefore, $Prob_{pred}(S_{t+1})$ can be calculated via \textbf{Equation \ref{eq:ProbDist_future}} where $\mathbf{P}_t^{a(r)}$ is one of the identified transition matrices which is correlated with $a(r)$. Only for the first prediction, the uniform distribution and the marginal transition matrix are used such that $Prob_{pred}(S_1)=\mathbf{P}_t^* \cdot \mathcal{U}(1, n_t)$. The Jensen-Shannon Divergence (JSD) method is implemented which can measure the difference between two probability distributions as described in \cite{lin1991divergence} for the comparison of the two probability distributions.

Under the same scenario and settings, the differences between the two probability distributions are quantified by the JSD method as shown in \textbf{Figure \ref{fig:result2}}. In the results, the difference between $Prob_{pred}(S_{1})$ and $Prob(S_{1})$ is relatively more significant than others in all cases because $Prob_{pred}(S_1)$ is calculated by using the uniform distribution and the marginal transition matrix. Except for the first prediction, the JSD values are less than 0.15 in all cases. Considering the JSD value is bounded 0 to 1, it is realized that e-FSM's prediction of the future state is accurate, claiming that transition-matrices are precisely identified through the proposed methodologies. In this study, the prediction of the one-step-ahead state is shown, but e-FSM can predict the further future state by using \textbf{Equation \ref{eq:ProbDist_future}} and \textbf{\ref{eq:ProbDist_future_multi}}.





%
\section{4. The overview of an online evolving framework}
\begin{figure}[!h]
  \centering
  \includegraphics[width=0.9\textwidth]{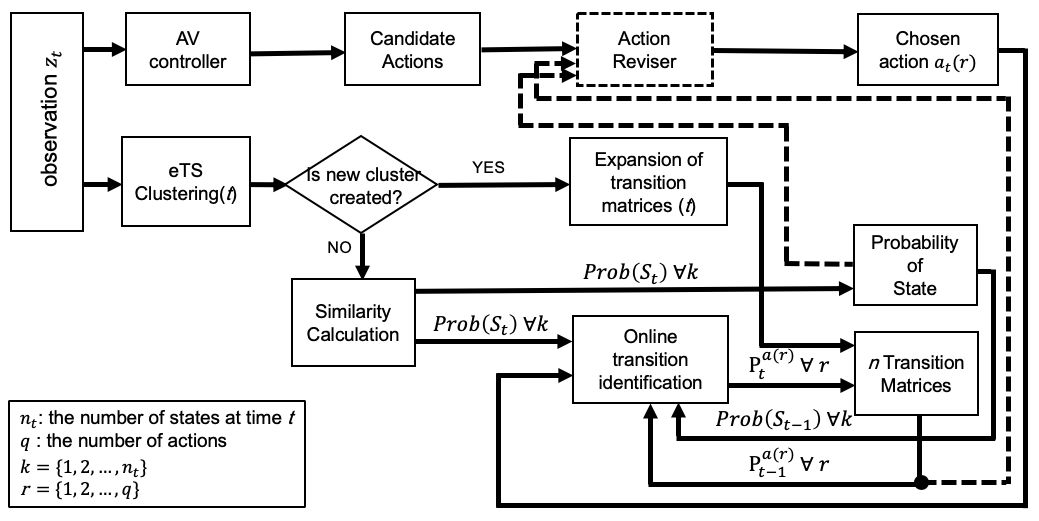}
  \caption{An online evolving framework for the safe automated vehicle control system}\label{fig:eFSM}
\end{figure}

An online evolving framework for safe AV control is proposed as shown in \textbf{Figure \ref{fig:eFSM}}. The framework is independent of the type of controller consisting of two sub-frameworks, an AV control framework and an e-FSM framework. 
In this study, specific steps in the $action$ $reviser$ module are not explained, but it is introduced how e-FSM assists the AV controller to choose a better action. In the framework, possible actions are returned by the independent decision-making of AV controller. The $action$ $reviser$ module either improves the returned actions or chooses the best action for safe AV control by using e-FSM's capabilities. As validated in the previous sections, e-FSM determines states uniquely and recognizes the state consistently, which can make the controller be able to detect initially unexpected dangerous situations. Also, e-FSM identifies state-transitions precisely so that the future states can be predicted accurately, which can support the controller to notice safer action for the future.

\section{5. Conclusion}
In this paper, specific properties and principles of e-FSM have been discussed, and its capabilities are validated under the simple car-following scenario. As shown in the experimental results, e-FSM can evolve its structure via the online state determination. The determined states represent unique situations, and the recognition of states is illustrated by the probability distributions. Through the proposed stochastic method, e-FSM identifies state-transitions precisely so that the accurate prediction of future states is possible. We claim that e-FSM can support the AV controller to determine unexpected situations, recognize states, and predict future states, which are required for better decision-making, in the online evolving framework.

\section{Acknowledgements}
This study is funded by the National Science Foundation (NSF) Cyber-Physical Systems (CPS) project under contract \#60046665. Authors would like to thank for the support.

\section{AUTHOR CONTRIBUTIONS}
The authors confirm contribution to the paper as follows: study conception and design:
T. Han, D. Filev, U. {\"O}zg{\"u}ner; data collection: T. Han; analysis and interpretation of results: T. Han, D. Filev, U. {\"O}zg{\"u}ner; draft manuscript preparation: T. Han, U. {\"O}zg{\"u}ner. All
authors reviewed the results and approved the final version of the manuscript.

\newpage
\nolinenumbers
\bibliographystyle{trb}
\bibliography{trb_template}

\end{document}